\documentclass{article}
\usepackage[colorlinks=true,
    linkcolor=red,
    urlcolor=blue,citecolor=blue]{hyperref}
\usepackage[preprint]{neurips_2023}
\usepackage{graphicx}
\usepackage[utf8]{inputenc} 
\usepackage[T1]{fontenc}    
\usepackage{hyperref}       
\usepackage{url}            
\usepackage{booktabs}      
\usepackage{amsfonts}        
\usepackage{nicefrac}        
\usepackage{microtype}      
\usepackage{xcolor}         
\usepackage{xspace}

\title{IQAGPT: Image Quality Assessment with Vision-language and ChatGPT Models\thanks{$\dagger$: Co-first authors; $\sharp$: Co-corresponding authors}}

\author{%
  Zhihao Chen$^\dagger$ \\
  ISTBI \\ Fudan University
  \And
  Bin Hu$^\dagger$ \\
  Huashan Hospital\\
  Fudan University
  \And
  Chuang Niu$^\dagger$ \\
  BME \& CBIS\\
  Rensselaer Polytechnic Institute
  \And
  Tao Chen \\
  ISTBI \\ Fudan University
  \AND
  Yuxin Li$^\sharp$ \\
  Huashan Hospital \\
  Fudan University
  \And
  Hongming Shan$^\sharp$ \\
  ISTBI \\ 
  Fudan University
  \And
  Ge Wang$^\sharp$ \\
  BME \& CBIS\\
  Rensselaer Polytechnic Institute
}

\begin{document}

\maketitle

\newcommand{\etal}{\textit{et al}.\xspace}
\newcommand{\ie}{\textit{i}.\textit{e}.\xspace}
\newcommand{\eg}{\textit{e}.\textit{g}.\xspace}
\newcommand{\tabincell}[1]{\begin{tabular}[l]{@{}l@{}} #1\end{tabular}}
\newcommand{\modelname}{IQAGPT\xspace}

\begin{abstract}
Large language models (LLMs), such as ChatGPT, have demonstrated impressive capabilities in various tasks and attracted an increasing interest as a natural language interface across many domains. 
Recently, large vision-language models (VLMs) like BLIP-2 and GPT-4 have been intensively investigated, which learn rich vision-language correlation from image-text pairs.
However, despite these developments, the application of LLMs and VLMs in image quality assessment (IQA), particularly in medical imaging, remains to be explored, which is valuable for objective performance evaluation and potential supplement or even replacement of radiologists' opinions. 
To this end, this paper introduces IQAGPT, an innovative image quality assessment system integrating an image quality captioning VLM with ChatGPT for generating quality scores and textual reports.
First, we build a CT-IQA dataset for training and evaluation, comprising 1,000 CT slices with diverse quality levels professionally annotated.
To better leverage the capabilities of LLMs, we convert annotated quality scores into semantically rich text descriptions using a prompt template. 
Second,  we fine-tune the image quality captioning VLM  on the CT-IQA dataset to generate quality descriptions. The captioning model fuses the image and text features through cross-modal attention.
Third, based on the quality descriptions, users can talk with ChatGPT to rate image quality scores or produce a radiological quality report.
Our preliminary results demonstrate the feasibility of assessing image quality with large models. Remarkably, our \modelname outperforms GPT-4 and CLIP-IQA, as well as the multi-task classification and regression models that solely rely on images.
\end{abstract}

\textbf{\textit{Keywords: }}Artificial intelligence,  computed tomography (CT), large language model, vision-language model, ChatGPT, GPT-4, image quality assessment, subjective evaluation

\section{Introduction}
In recent years, there have been many advances in the field of large language models (LLMs), such as PaLM~\cite{palm}, LLaMA~\cite{touvron2023llama} and GPTs~\cite{GPT-1,GPT-2,GPT-3}, which have shown excellent results in natural language processing (NLP) tasks including language translation, question answering, and text generation.  The most remarkable breakthrough is ChatGPT, which is built upon InstructGPT~\cite{InstructGPT} using labeler-written prompts and reinforcement learning from human feedback~\cite{RLHF}. 
As a result, ChatGPT can produce more informative and human-like responses and interact quite naturally with users in a conversational manner, allowing it to help humans accomplish many tasks, such as writing and programming~\cite{chen2021evaluating}.

However, LLMs such as ChatGPT are unable to cope with visual information since they are only trained on textual data. To address this gap, visual-language models (VLMs)~\cite{beit-3,blip2,palm-e,visual_chatgpt,park2023selfsupervised, niu2023ct,lyu2023translating}, which synergistically combine the capabilities of LLMs with visual processing, were proposed to capture rich vision-language correspondence and perform well in various multi-modal tasks such as report generation, diagnosis, and vision question answering. 
Recently, attempts have been made to interface LLMs, such as ChatGPT, with VLMs to understand visual information in computer vision tasks, leading to the development of multimodal models.
In this context, OpenAI launched its new large vision-language model, GPT-4~\cite{openai2023gpt4}, with amazing results on multimodal tasks during dialogues.
Wu~\etal~\cite{visual_chatgpt} build a system, called Visual ChatGPT, by incorporating different vision models for users to interact with ChatGPT beyond language input. 
MiniGPT-4~\cite{zhu2023minigpt} integrates an advanced large language model (LLM), Vicuna~\cite{chiang2023vicuna}, and a pre-trained ViT~\cite{vit} with a single linear projection layer, leading to a performance close to that of GPT-4.

\begin{figure}[t]
\centering
\includegraphics[width=0.8\linewidth]{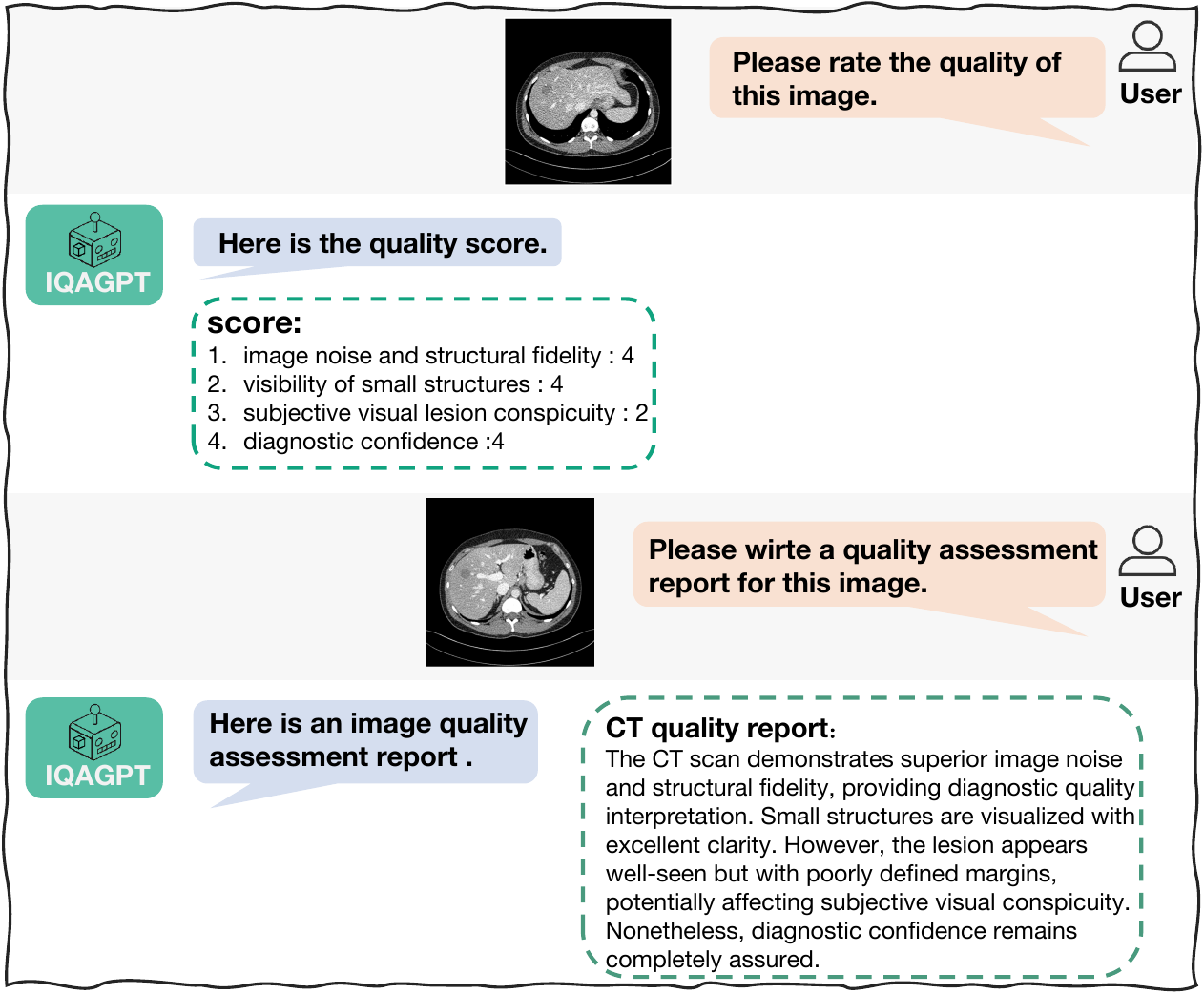}
\caption{A dialogue between humans and the proposed \modelname. In the dialogues, \modelname can output scores and write the quality report based on an input image}
\label{example}
\end{figure}

While LLMs and VLMs are powerful in many tasks, up to now few efforts have been made to adapt them for image quality assessment (IQA), which plays an important role in imaging and image processing applications such as image restoration~\cite{sarmah2023survey, pack2022cardiac,lei2023ct} and medical diagnosis~\cite{niu2023advances, al2023vision}. 
Particularly, in the field of computed tomography (CT), reconstructed low-dose CT images from various deep learning methods~\cite{redcnn,wgan-vgg,shan20183,map-nn,fu2022deep,corediff,chen2023ascon,chen2023lit} may lead to blurring or over-smoothing problems, hindering their clinical translation. 
Therefore, assessing CT image quality prior to diagnosis is essential.
Over the past decades, several objective IQA metrics have been widely used, including peak signal-to-noise ratio (PSNR), structural similarity (SSIM), and root-mean-square error (RMSE). However, these metrics are neither well aligned with the subjective evaluations of radiologists nor with diagnostic accuracy. 

Consequently, developing a medical image quality assessment system is highly desirable for radiologists to assess clinically relevant image quality such as lesion conspicuity. 
The most effective way is to invite experienced radiologists to review images~\cite{singh2010abdominal} and give ground truth labels. However, this method is expensive and time-consuming. 
While several deep learning-based image quality assessment methods exist, they primarily focus on natural images with little linguistic interactivity.~\cite{madhusudana2022image2,madhusudana2022conviqt,gao2019blind}. 
CLIP-IQA~\cite{clipiqa} utilizes a frozen CLIP model to calculate the similarity between images and predefined positive and negative prompts. Nevertheless, it was intended for natural images and can only train on simple text prompts, one metric each time. These limitations make it inadequate for complex medical IQA, particularly for evaluating small structures and lesions in CT images.
An interesting question here we ask is: \emph{Could we build a ChatGPT-like system based on VLMs to mimic the image quality evaluation by radiologists?} 

In this paper, we propose an image quality assessment system with VLMs and ChatGPT, termed as \modelname. 
We build \modelname based on an image quality captioning VLM and incorporate it with ChatGPT to generate quality scores and summarize a quality report of CT images to be assessed.
First, to train our \modelname, we collect a dataset of 1,000 image-text pairs, named CT-IQA, in which an experienced radiologist was asked to score CT images of different qualities similar to the subjective evaluation previously reported~\cite{singh2010abdominal,map-nn}, including image noise, small structures, lesion conspicuity, and diagnostic confidence.
To utilize the strengths of LLMs in subjective image evaluation, we design a prompt template to convert the quality scores to text descriptions. 
Second, we develop an image quality captioning model built upon a pre-trained medical VLM~\cite{park2023selfsupervised} and fine-tune it on the CT-IQA dataset with an autoregressive language modeling objective that predicts the next token given the previous tokens~\cite{GPT-1}.
Finally, through interacting with ChatGPT, \modelname can score CT images and generate quality reports based on the caption from the image quality captioning model. 
Figure~\ref{example} presents an exemplary dialogue between a user and the proposed \modelname.

In summary, the main contributions of this work are as follows.
\begin{itemize}
    \item We introduce a hybrid large model approach for CT image quality assessment, which synergizes the objective and subjective image quality evaluation in a clinically important scenario.
    \item Specifically, we present an image quality assessment system
    consisting of VLMs and ChatGPT, termed \modelname, which is built on an image quality captioning model and can output quality scores and reports by interacting with ChatGPT.
    \item We collect a CT-IQA dataset for IQA, containing 1,000 image-text pairs professionally annotated according to four common subjective metrics used in diagnosis.
    \item Preliminary results demonstrate the feasibility of assessing CT image quality with \modelname, and the resulting text-guided image quality captioning model outperforms GPT-4 and CLIP-IQA.
\end{itemize}
\section{Methods and Materials}
\label{method}
The goal of this study is to develop an image quality assessment system with VLMs and ChatGPT, called \modelname. 
In the following, we ﬁrst report on the CT-IQA dataset in Subsection~\ref{CT_IQA}. Then, in Subsections~\ref{caption_model} and~\ref{ChaGPT}, we describe the image quality captioning model and our \modelname that interacts with ChatGPT, respectively.  Furthermore, we give the implementation details in Subsection~\ref{implementation} and explain how we evaluate the performance of \modelname in Subsection~\ref{evaluation}.

\begin{figure}[htbp]
\centering
\includegraphics[width=1\linewidth]{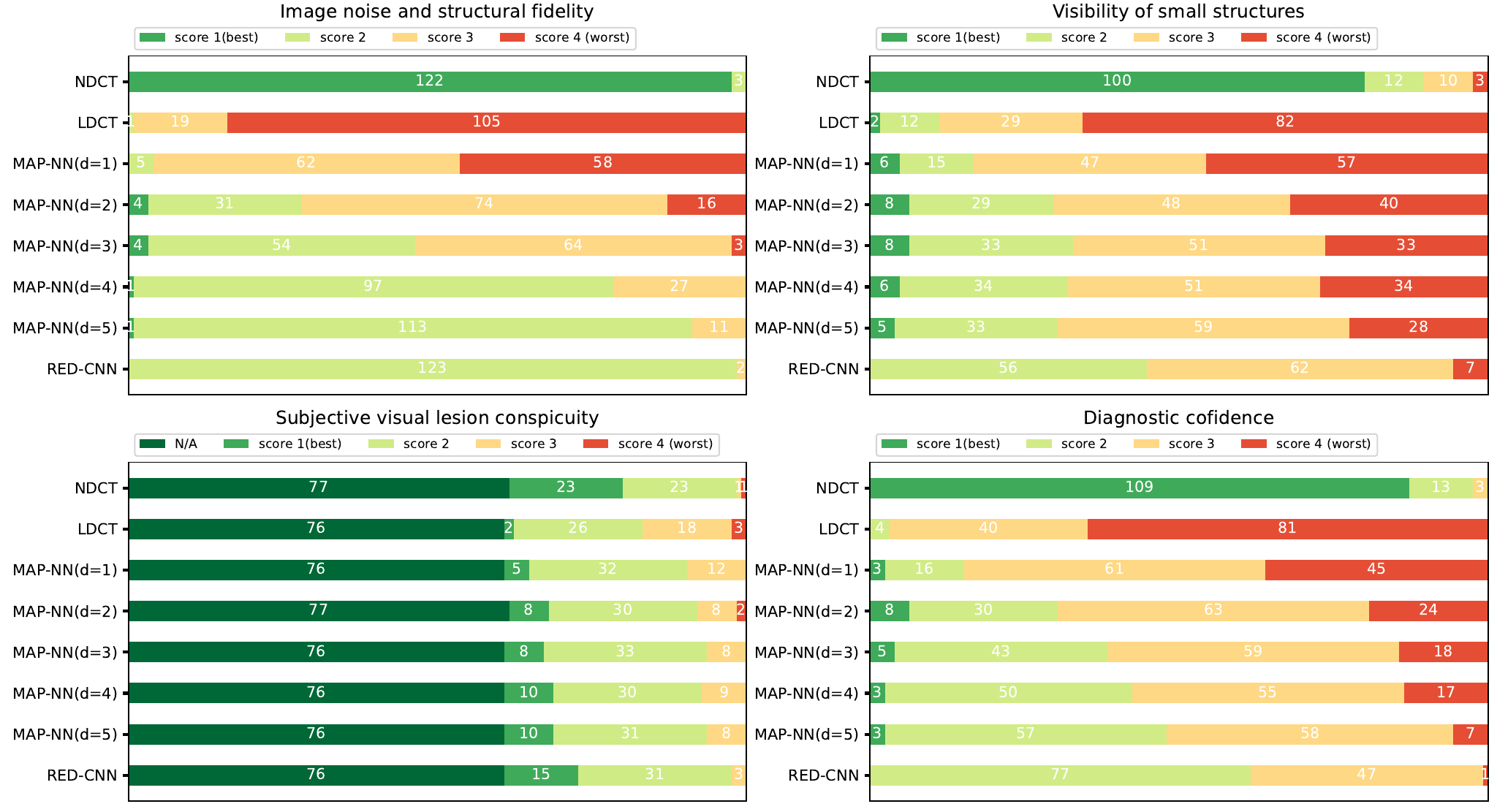}
\caption{The distribution of scores of the four metrics assessed by the radiologist in constructing our CT-IQA dataset. Scores 1, 2, 3, and 4 are defined in Subsection~\ref{CT_IQA}.}
\label{dataset_freq1}
\end{figure}

\subsection{CT-IQA Dataset}
\label{CT_IQA}
To adapt to the IQA task and achieve an accurate quality assessment of CT images, we created an image-text dataset called CT-IQA, in which an experienced radiologist was asked to assess CT images subjectively. 
To this end, we randomly selected normal-dose CT (NDCT) slices and corresponding low-dose CT images (LDCT) at 25$\%$ of the normal dose from the 2016 AAPM Grand Challenge dataset~\cite{mayo2016}, which includes abdominal CT scans of 10 anonymous patients. 
Specifically, we selected 100 NDCT and LDCT pairs uniformly from 8 patients for training and 25 slice pairs uniformly from the remaining 2 patients for testing.
Additionally, we randomly simulated some lesions in NDCT and corresponding LDCT to evaluate subjective visual lesion conspicuity. 
Next, we processed the selected 125 LDCT images with a modularized denoising model~\cite{map-nn} called MAP-NN, which produced various intermediate denoised images with associated noise reduction directions.
Besides, we implemented RED-CNN~\cite{redcnn}, a widely-used denoising model that was optimized with the MSE loss function. 
Finally, we obtained 1,000 CT slices with different quality, including 125 NDCT slices with corresponding 125 LDCT slices, 625 reconstructed images with 5 denoising levels from MAP-NN, and 125 reconstructed images from RED-CNN.
We employed the abdomen window of all CT scans [-160,240] HU for  visualization of abdominal organs. 
The radiologist was asked to score these CT images in terms of four metrics used in the previous study~\cite{singh2010abdominal,map-nn}, which are defined as follows:
\begin{itemize}
\item Image noise and structural fidelity on a four-point scale
(1 = Better than usual, acceptable for diagnostic interpretation; 2 = Average, acceptable for diagnostic interpretation; 3 = Sub-optimal, for limited diagnostic information only, 4= Unacceptable for diagnostic interpretation)
\item The visibility of small structures (small blood vessels, adrenal glands, small lymph nodes)  on a four-point scale
(1 = excellent visualization, 2 = acceptable visibility, 3 = sub-optimal visibility, and 4 = unacceptable visualization)
\item Subjective visual lesion conspicuity (N/A = if no lesion ) on a four-point scale
(1 = well-seen lesion with well-visualized margins, 2 = well-seen lesion with poorly visualized margins, 3 = poorly seen lesion with poorly visualized margins, and 4 = lesion blurred with severe loss of margins)
\item Diagnostic confidence on a four-point scale
 (1 = completely confident; 2 = probably confident; 3 = confident only for a limited clinical entity such as a kidney stone, a calcified lesion,  or a large lesion; and 4 = poor confidence).
\end{itemize}
Figure~\ref{dataset_freq1} shows the distribution of
human expert scores across the aforementioned four metrics for CT images of 8 different image qualities, including NDCT, LDCT, MAP-NN (d=1), MAP-NN (d=2), MAP-NN (d=3), MAP-NN (d=4), MAP-NN (d=5), and RED-CNN. $d$ represents the denoising level.

\begin{figure}[htbp]
\centering
\includegraphics[width=1\linewidth]{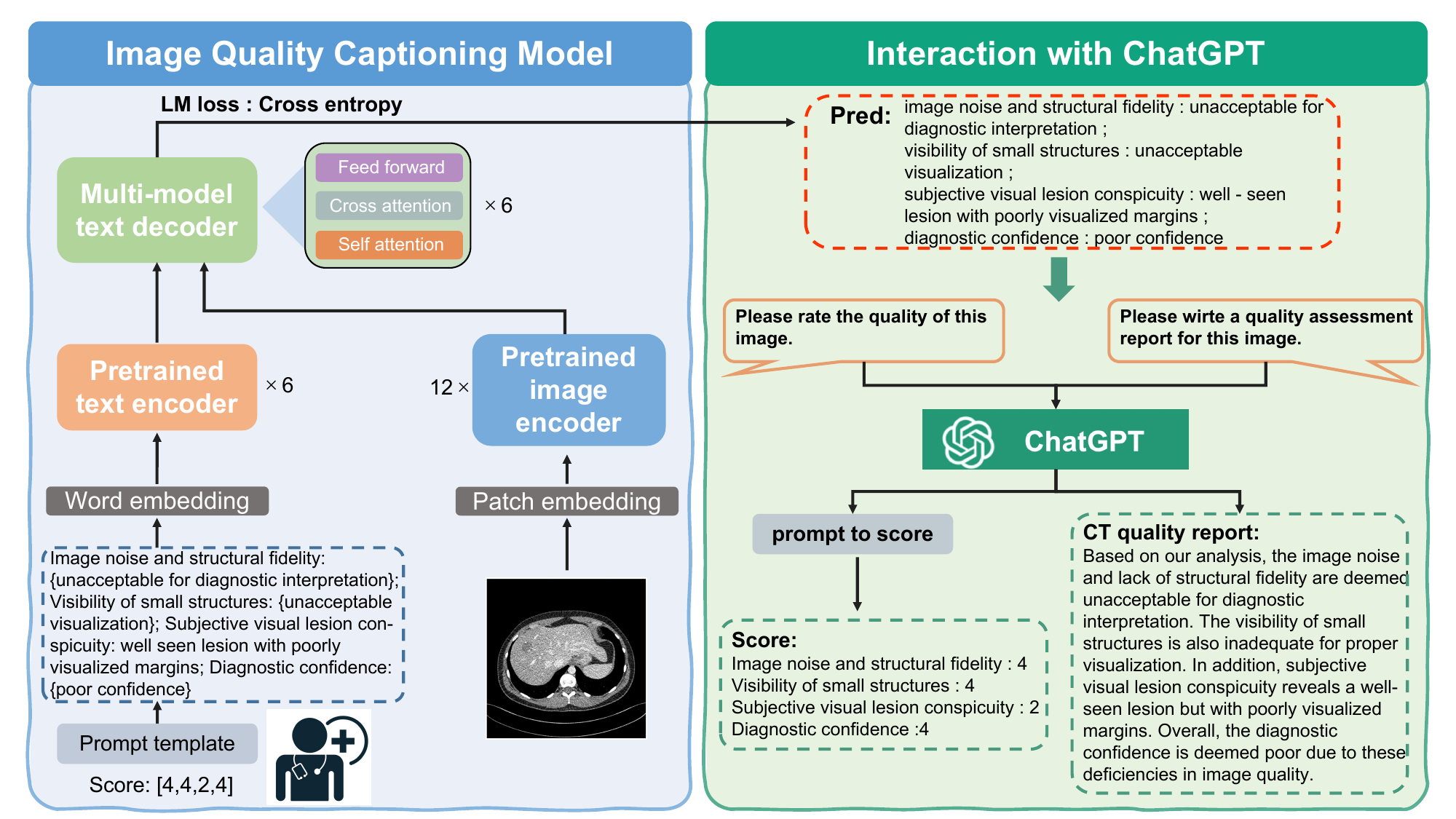}
\caption{Overview of \modelname. While the left side shows our image quality captioning model, the right side details the process of the score and report generation through interacting with ChatGPT.}
\label{IQA_net}
\end{figure}

\subsection{Image Quality Captioning Model}
\label{caption_model}
Instead of using the rating scores to train a classification or regression model, we develop an image quality captioning model to summarize the image quality.
By doing so, the vision language model with semantic text information and image-text fusion can better appreciate the subjective scores than image-only models, which will be further discussed in Section~\ref{experiment}. 
Our model is based on a pre-trained medical VLM and fined-tuned with an autoregressive language modeling objective on the CT-IQA dataset. 
To leverage the capabilities of LLMs in the subjective image evaluation, we convert scores to quality descriptions using a specific prompt template during training.
Our prompt template is defined as that 
``Image noise and structural fidelity: $\{$description 1$\}$; Visibility of small structures: $\{$description 2$\}$; Subjective visual lesion conspicuity: $\{$description 3$\}$; Diagnostic confidence: $\{$description 4$\}$''. 
Every description is the evaluation criterion corresponding to the score described in Subsection~\ref{CT_IQA}. 
An example to convert the score to the quality caption is given in the lower left part of Figure~\ref{IQA_net}, where the score assessed by the radiologist is [4,4,2,4]. 

Our image quality captioning model consists of an image encoder, a text encoder, and a multi-modal text decoder, as shown on the left side of Figure~\ref{IQA_net}. We use a 12-layer visual transformer ViT-S/16~\cite{vit} as the image encoder and the first 6 layers of the $\mathrm{BERT}_\mathrm{base}$~\cite{bert} model as the text encoder. The multi-modal text decoder is the last 6 layers of the $\mathrm{BERT}_\mathrm{base}$ to fuse image and the text features through cross-modal attention. 
The image encoder, text encoder, and multimodal decoder were pre-trained in radiography images and report pairs~\cite{park2023selfsupervised} using four learning objectives: contrastive learning for cross- and intra-modal alignment, masked language modeling for image-guided text completion, masked image modeling for text-guided image completion, and image-text matching; please refer to~\cite{park2023selfsupervised} for more details on these four objectives. 

We fine-tuned the pre-trained models with the next word prediction to allow the auto-regressive generation of image captions for CT image quality assessment. We hypothesize that this paradigm, combined with our input template, allows LLMs to better comprehend the relationship between different metrics.
We denote a CT-text pair as $(I,T)$, where $I$ represents a CT slice and $T$ is defined as $T=\left(t_{1}, t_{2}, \ldots, t_{m}\right)$ with $m$ tokens.  
The objective is to maximize the following log-likelihood:
\begin{equation}
\mathcal{L}\left(I,T\right)=\sum_{i=1}^{m} \log P\left(t_{i} \mid t_{1}:t_{i-1},I; \theta\right),
\end{equation}
where $P$ is the conditional probability modeled by the image quality captioning model, and $\theta$ represents the trainable parameters of the model.

\subsection{Interaction with ChatGPT}
\label{ChaGPT}
ChatGPT provides a language interface with remarkable reasoning capabilities across many domains~\cite{visual_chatgpt}. In our \modelname, we enable an interaction between ChatGPT and users, aimed at generating more comprehensive output information, as depicted on the right side of Figure~\ref{IQA_net}.
When a user uploads a CT image, they can prompt \modelname with a request like ``Please rate the quality of this image.'' or ``Please write a quality assessment report for this image.''
Subsequently, the user receives either a quality score or a detailed quality report.
To this end, we use ChatGPT to perform corresponding operations on the output caption from the image quality captioning model. 
For the score-related demands, it converts the predicted caption to the score according to the prompt template described in Subsection~\ref{CT_IQA}.
For the report-related demands, it summarizes the predicted caption into a quality assessment report in a radiology report format. 

\subsection{Implementation Details}
\label{implementation}
We trained our models with NVIDIA V100 GPUs. In the training process, we fine-tuned the image quality captioning model in \modelname for 50 epochs based on the pre-trained model~\cite{park2023selfsupervised}, in which we used the AdamW optimizer~\cite{adamw} and the weight decay of 0.02. 
We initialized the learning rate at $2.0\times 10^{-4}$, warm-up~\cite{warmup} in the first 2 epochs with the learning rate of $1.0\times 10^{-5}$, and gradually reduced it to $1.0\times 10^{-6}$ with the cosine annealing~\cite{conisneLR}. 
For data processing, we employed the full-size images within an abdomen window of [-160, 240] HU and split 10 patients into the training and testing datasets by 8:2 as described in 
Subsection~\ref{CT_IQA}.
Also, we randomly augmented training samples using horizontal ﬂipping and rotation. 

\subsection{Evaluation Metrics}
\label{evaluation}
To show the effectiveness of \modelname, we 
quantitatively evaluated the performance of generated quality captioning and scores. First, we analyzed captioning results utilizing widely recognized metrics in text generation tasks:
Bilingual Evaluation Understudy (BLEU-n; ``n'' means $n$ words)~\cite{bleu}, Recall Oriented Understudy of Gisting Evaluation (ROUGE-L; ``L'' means the longest common subsequence)~\cite{rouge}, Metric for Evaluation of Translation with Explicit Ordering (METEOR)~\cite{meteor}, and Consensus-based Image Description Evaluation (CIDEr; ``r'' stands for recall)~\cite{cider}.
These metrics measure the similarity between the generated and reference texts, with higher scores for better quality.
Specifically, BLEU measures the quality of machine-translated text compared to a human reference translation. It computes an  $n$-grams (phrases of $n$ words) based precision for the candidate sentence with respect to the references.
ROUGE-L focuses on the longest common subsequence between the evaluated text and the reference text.
METEOR is based on the harmonic mean of unigram precision and recall, with recall weighted higher than precision.
CIDEr measures the similarity of a generated sentence to a set of reference sentences by considering human consensus.
Notably, BLEU-n, ROUGE-L, and METEOR scores range from 0 to 1 while CIDEr ranges from 0 to infinity. In addition, we converted the output text descriptions into scores, compared the performance in terms of accuracy as the classification evaluation, and computed the Pearson linear correlation coefficient (PLCC) and Spearman’s rank order correlation coefficient (SROCC) as the regression evaluation.

\section{Results}
\label{experiment}

\begin{figure}[htbp]
\centering
\includegraphics[width=1\linewidth]{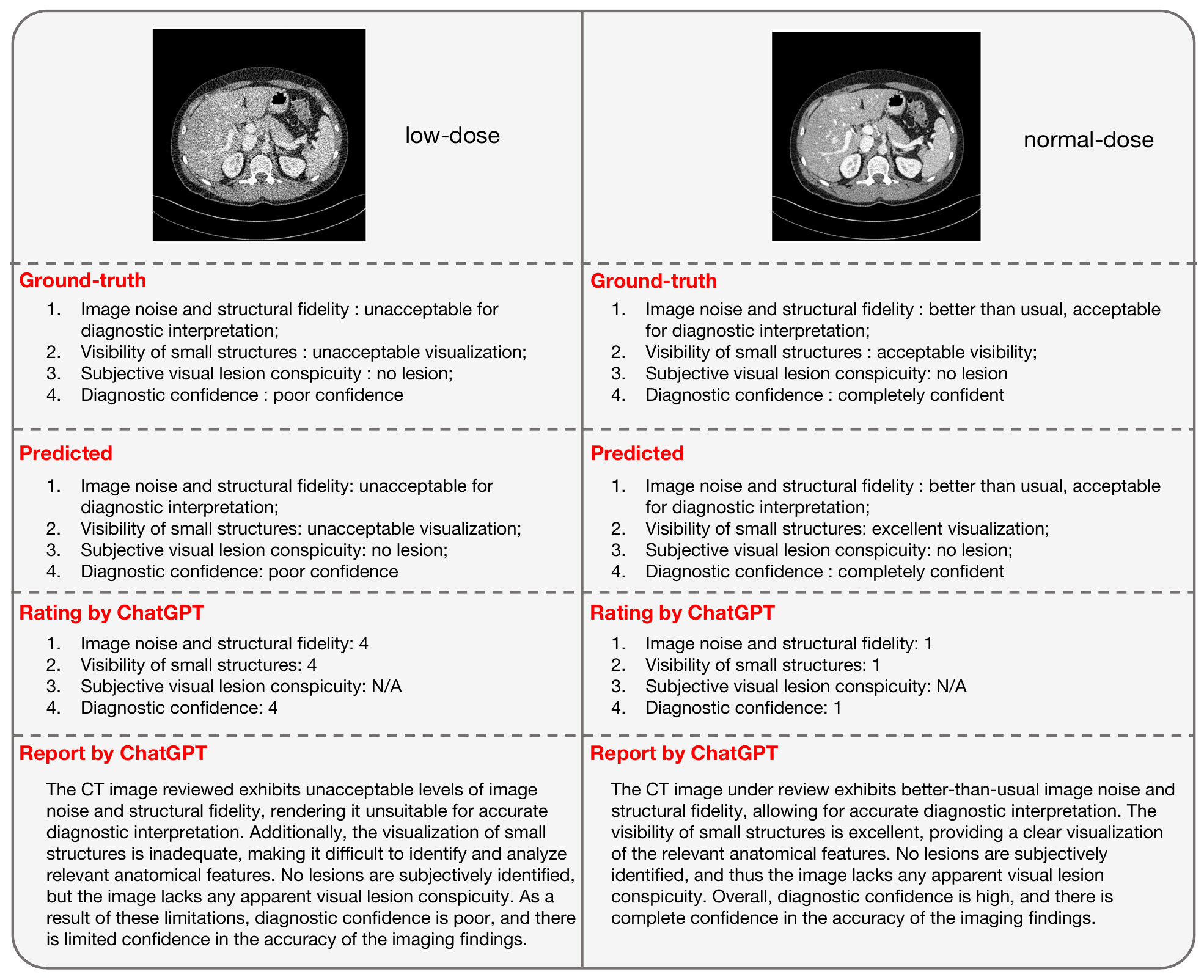}
\caption{The captions predicted using our method and ChatGPT’s generated scores and reports.}
\label{caption_example1}
\end{figure}

\subsection{Evaluation of Generated Quality Captioning}
\begin{table}[htbp]
\centering
\caption{Quantitative evaluation of captioning quality generated using \modelname and MiniGPT-4 respectively.}
\label{caption_result}
\resizebox{\textwidth}{!}{
\begin{tabular}{lccccccc}
\toprule[1.5pt]
Method  & BLEU-1&  BLEU-2&  BLEU-3	& BLEU-4 &METEOR&	ROUGE-L&	CIDEr\\
 \midrule
MiniGPT-4 &  0.798	&0.733 &	0.717&	0.652	&0.516	&0.826&	3.070 \\
\modelname  & \textbf{0.819} &\textbf{0.777} &	\textbf{0.742} & \textbf{0.712}	& \textbf{0.546}&\textbf{0.858}&	\textbf{3.620} \\
\bottomrule[1.5pt]
\end{tabular}}
\end{table}
In Figure~\ref{caption_example1}, we present two examples of our test results, where we converted the predicted descriptions to scores and quality reports using ChatGPT. 
It can be observed that \modelname consistently generates quality descriptions in excellent alignment with radiologists' annotations.
Furthermore, the reports generated using ChatGPT are consistent with the outputs from our quality captioning model, which effectively overcomes the limitations of the existing VLM dialogue when assessing the quality of medical images.
Then, we compared the quantitative captioning performance of \modelname and MiniGPT-4, as depicted in Table~\ref{caption_result}.
We did not employ GPT-4~\cite{openai2023gpt4}, as its latest version, GPT-4V, is not tailored for interpreting specialized medical imagery such as CT scans. 
We fine-tuned the learnable linear layer in MiniGPT-4 using our CT-IQA dataset in their experimental settings~\cite{zhu2023minigpt}.
\modelname achieves better quantitative results in seven metrics because the large language model (Vicuna~\cite{chiang2023vicuna}) in MiniGPT-4 lacks the expertise of CT quality assessment and it was frozen during training. 
Consequently, while MiniGPT-4 was instable, our \modelname produced the quality description very consistent with radiologists' annotations. 

\begin{figure}[htbp]
  \centering
    \includegraphics[width=0.85\linewidth]{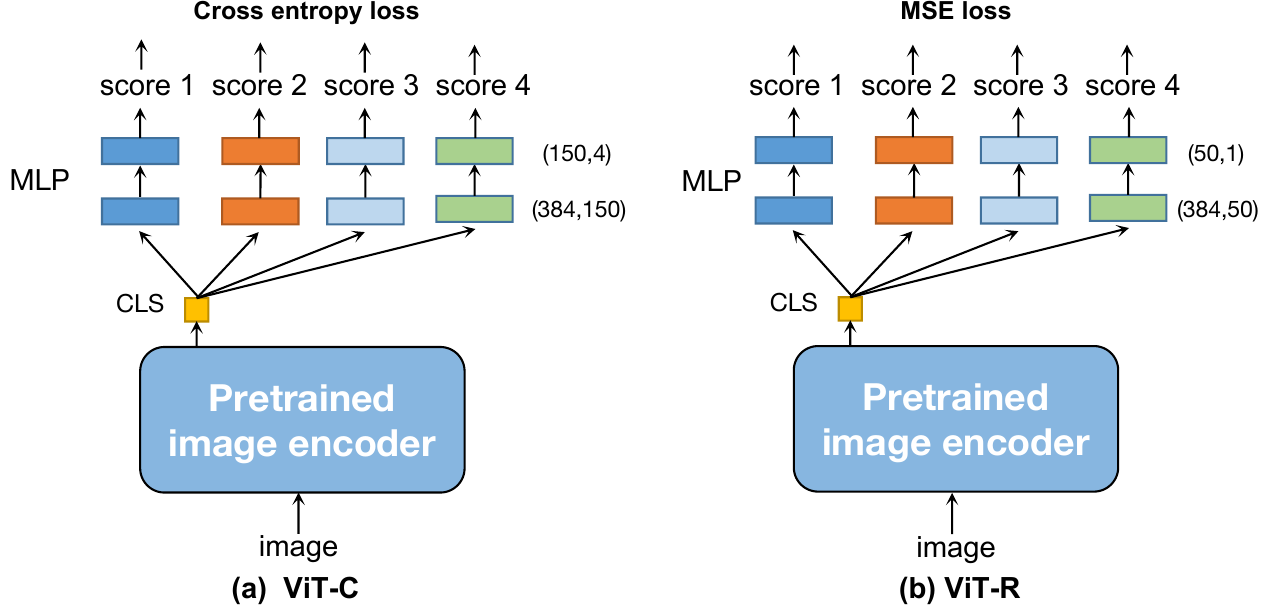}
    \caption{The flowcharts of (a) the multi-task classification model ViT-C and (b) the multi-task regression model ViT-R, respectively. CLS Tokens are followed by four groups of classifiers, each consisting of two fully-connected layers. Scores 1, 2, 3, and 4 are the categories corresponding to the four metrics described in Subsection~\ref{CT_IQA}.}
    \label{mtc_mtr_net}
\end{figure}

\subsection{Evaluation of Generated Quality Score}
To validate the efficacy of our image quality captioning model, we  employed our prompt template that transforms output text descriptions into scores for assessing \modelname's performance in both the classification and regression tasks.
Specifically, we conducted a comparative study on \modelname with an image-only multi-task classification model, using accuracy as a metric. Additionally, \modelname was compared against CLIP-IQA~\cite{clipiqa} and an image-only multi-task regression model, employing PLCC and SROCC. These quantitative results are detailed in Tables~\ref{compare_class} and~\ref{compare_regre}. 
It is pertinent to note that CLIP-IQA+ represents the fine-tuned version of CLIP-IQA.
The calculation of PLCC and SROCC for the metric of subjective visual lesion conspicuity was not performed, as over half of the CT scans in our dataset do not contain lesions.
For the image-only multi-task classification and regression models, we name them ViT-C and ViT-R respectively. First, we utilized the same pre-trained image encoder (ViT-S/16) in \modelname to extract image features. Then, four pairs of fully-connected layers were implemented following the classification (CLS) token for four metrics, as depicted in Figure~\ref{mtc_mtr_net}. The ViT-C and ViT-R employed cross-entropy loss and mean squared error loss respectively. We used the same training strategy with \modelname for training CLIP-IQA+, ViT-C, and ViT-R.

\begin{table}[htbp]
    \centering
    \caption{Performance comparison of \modelname with ViT-C in the classification evaluation in terms of accuracy.}
    \begin{tabular*}{0.98\linewidth}{@{\extracolsep{\fill}}ccc}
    \toprule[1.5pt]
     & ViT-C  & \modelname  \\
   
    \midrule
    Image noise and structural fidelity &
   0.545 & \textbf{0.765}    \\
    
     Visibility of small structures & 
   0.405 & \textbf{0.620}   \\
     
     Subjective visual lesion conspicuity &
   0.725 &  \textbf{0.820}     \\
     
     Diagnostic confidence  & 0.375 & \textbf{0.605}    \\
     
     Mean  & 0.512  &\textbf{0.702}   \\
    \bottomrule[1.5pt]
    \end{tabular*}
    \label{compare_class}
\end{table}

\begin{table}[htbp]
    \centering
    \caption{Performance comparison of \modelname with CLIP-IQA and ViT-R in the regression evaluation in terms of PLCC and SROCC (PLCC/SROCC).}
    \resizebox{\textwidth}{!}{
    \begin{tabular}{cccccc}
    \toprule[1.5pt]
     & CLIP-IQA & CLIP-IQA+ & ViT-R  & \modelname\\
   
    \midrule
    Image noise and structural fidelity  & 0.277/0.271 & 0.742/0.633  &0.580/0.460 &
     \textbf{0.821/0.820}  \\
    
     Visibility of small structures  & 0.121/0.117 & 0.712/0.696 & 0.436/0.415 & 
      \textbf{0.743/0.735}\\
     
     Subjective visual lesion conspicuity &
      \textbf{-}&\textbf{-}&\textbf{-}&\textbf{-}\\
     
     Diagnostic confidence   & 0.081/0.069 &  0.650/0.642 &0.504/0.422 &\textbf{0.699/0.689}\\
     
     Mean   & 0.160/0.114 & 0.701/0.657   &0.531/0.519 & \textbf{0.754/0.748} \\
    \bottomrule[1.5pt]
    \end{tabular}}
    \label{compare_regre}
\end{table}

In Table~\ref{compare_class}, it is evident that \modelname outperformed the image-only classification model ViT-C across four metrics, achieving a notable improvement of 0.19 in mean accuracy. For regression, \modelname surpassed CLIP-IQA, as shown tasks in Table~\ref{compare_regre}. 
In addition, a notable advantage of \modelname is its efficiency; unlike CLIP-IQA, which requires separate fine-tuning for each of the four metrics, \modelname is capable of simultaneously producing results for all metrics as a single output.

\begin{table}[htbp]
    \centering
    \renewcommand\arraystretch{1.1}
    \caption{Accuracy for each of four metrics in eight image quality levels. Metric 1: Image noise and structural fidelity; Metric 2: Visibility of small structures; Metric 3: Subjective visual lesion conspicuity; and Metric 4: Diagnostic conﬁdence. MAP-NN$(\cdot)$ provides 5 denoising levels~\cite{map-nn}.}
    \label{accuracy_every}
    \resizebox{\textwidth}{!}{
    \begin{tabular}{lccccccccc}
    \toprule[1.5pt]
         & NDCT & LDCT & MAP-NN(1) & MAP-NN(2) & MAP-NN(3) & MAP-NN(4) & MAP-NN(5) & RED-CNN & Mean  \\ 
         \midrule
        Metric 1 & 1.000  & 0.800  & 0.600  & 0.520  & 0.480  & 0.880  & 0.880  & 1.000  & 0.765   \\ 
        Metric 2 & 0.960  & 0.680  & 0.480  & 0.360  & 0.520  & 0.600  & 0.640  & 0.720  & 0.620   \\ 
        Metric 3 & 0.920  & 0.920  & 0.880  & 0.760  & 0.800  & 0.840  & 0.760  & 0.800  & 0.820   \\ 
        Metric 4 & 0.920  & 0.760  & 0.360  & 0.400  & 0.320  & 0.440  & 0.680  & 0.960  & 0.605   \\ 
        Mean & 0.950  & 0.790  & 0.580  & 0.510  & 0.520  & 0.690  & 0.740  & 0.840  & 0.702  \\ 
        \bottomrule[1.5pt]
    \end{tabular}}
\end{table}

For each image quality level and metric, we computed accuracy using converted scores, as depicted in Table~\ref{accuracy_every}. 
The relative accuracies associated with intermediate images generated by MAP-NN may not be highly robust, due to their similar image features. This aligns with the challenges in the subjective evaluation of images with subtle quality differences, a critical aspect of our CT-IQA dataset. Our study highlights the complexity of differentiating between similar images.

\begin{figure}[htbp]
\centering
\includegraphics[width=1\linewidth]{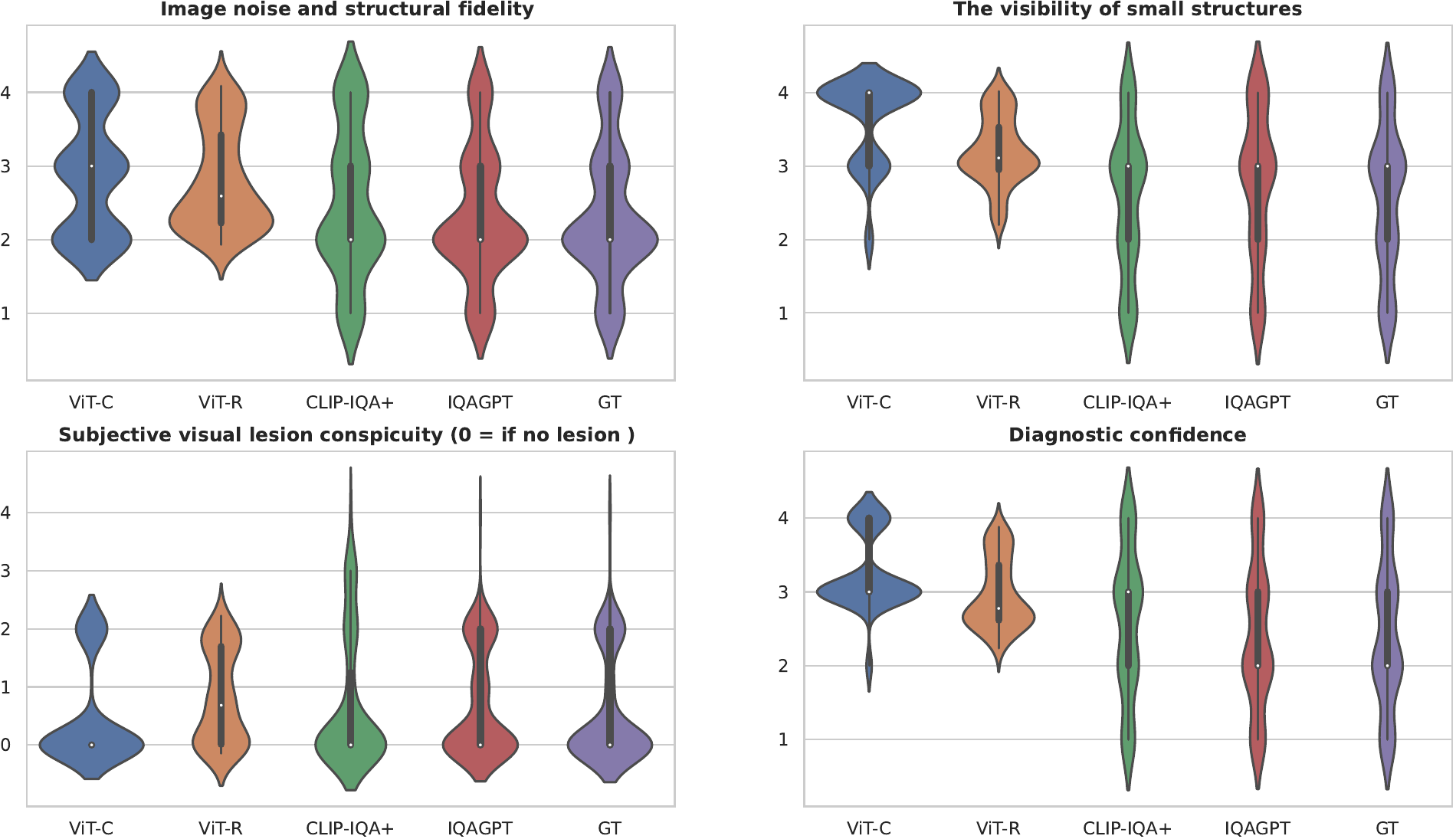}
\caption{Scores distribution for four quality metrics using \modelname, ViT-C, ViT-R, and CLIP-IQA+. The last column lists the ground-truth (GT) scores. }
\label{score_distribution}
\end{figure}

Furthermore, we present the score distributions of  \modelname, ViT-C, ViT-R, and CLIP-IQA+ for four quality metrics in Figure~\ref{score_distribution}.
Notably, our method more closely approximates the ground truth compared to ViT-C, ViT-R, and CLIP-IQA+, demonstrating its effectiveness. 
Overall, our method has a higher correlation with human perception than the competing methods, marking a significant advancement in CT subjective image quality assessment.

\begin{figure}[htbp]
\centering
\includegraphics[width=0.9\linewidth]{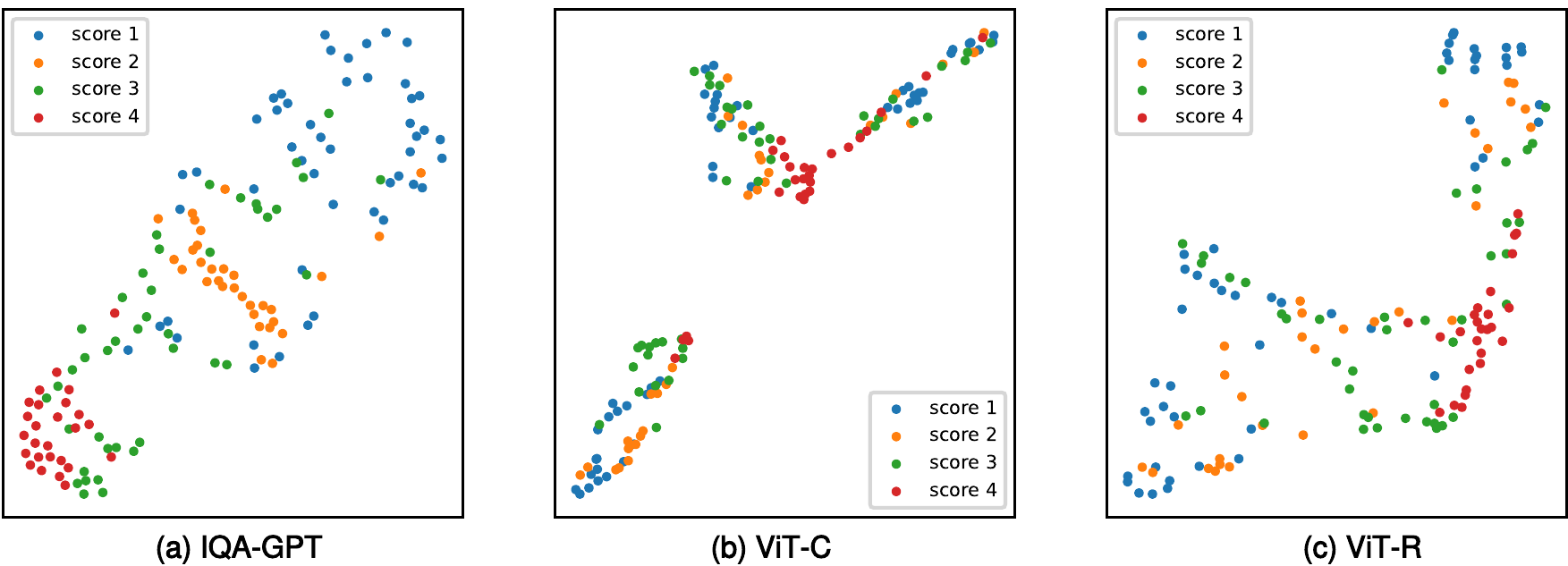}
\caption{Feature visualization of the CLS token in the image encoder of (a) \modelname, (b) ViT-C, and (c) ViT-R, using the t-SNE method. The samples are labeled with categories from the metric of ``Image noise and structural fidelity''.}
\label{cls_tsne}
\end{figure}

\begin{figure}[htbp]
\centering
\includegraphics[width=0.6\linewidth]{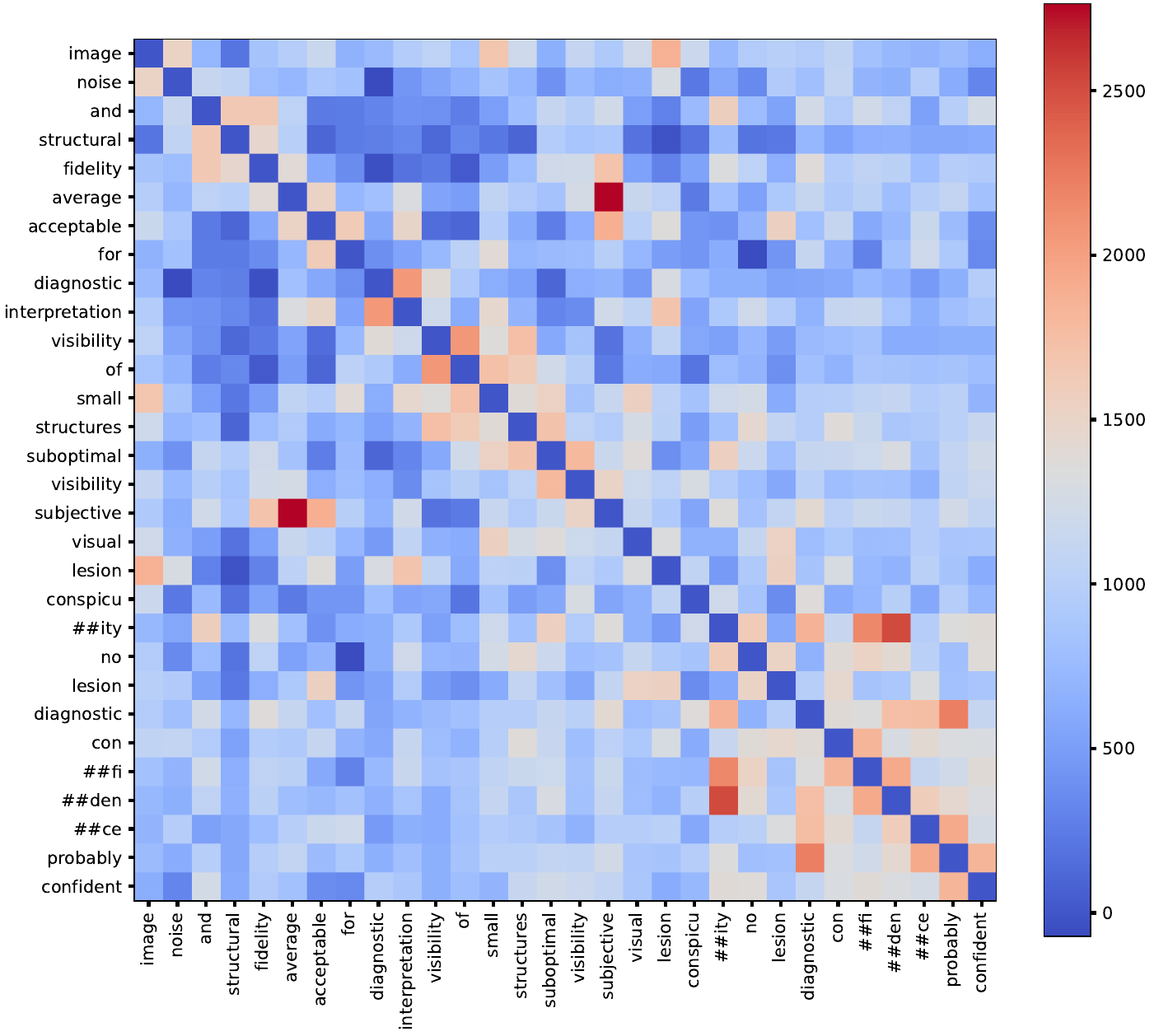}
\caption{The self-attention map of tokens from the last layer in the multi-modal text decoder.}
\label{text_decoder_attn}
\end{figure}

\subsection{Ablation on LLMs}

To further demonstrate the effectiveness of textual semantic information, we employed the $t$-SNE~\cite{tsne} method to visualize the features of the CLS token in the image encoders of \modelname, ViT-C, and ViT-R, as illustrated in Figure~\ref{cls_tsne}. Each sample was labeled using the score of the  ``Image noise and structural fidelity'' metric. This visualization demonstrates that \modelname distinguished features of different categories more clearly than ViT-C and ViT-R, and exhibiting an ordered sequence in the score-based feature representation.  
Additionally, the self-attention map of tokens from the multi-modal text decoder, depicted in Figure~\ref{text_decoder_attn}, reveals that each token is interconnected not only with tokens from the same task but also with those from preceding tasks. This finding underscores the merits of textual descriptions in capturing inter-task correlations, thereby enhancing the classification performance.

\begin{figure}[htbp]
\centering
\includegraphics[width=1\linewidth]{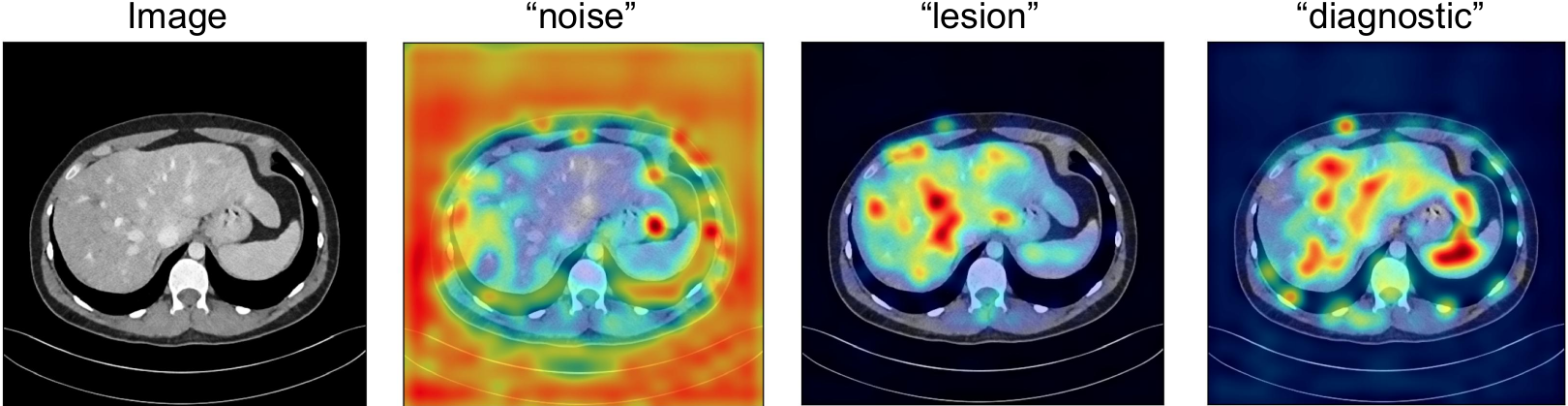}
\caption{Grad-CAM visualizations on the cross-attention maps corresponding to individual words.}
\label{grad_cam}
\end{figure}

\subsection{Interpretation}
To provide an interpretation of our quality captioning model, we present the per-word Grad-CAM visualizations  in Figure~\ref{grad_cam}. It can be seen that the Grad-CAM visualizations are highly correlated with where radiologists would look at when making decisions. For instance, radiologists tend to concentrate on the global appearance of an image when assessing ``\emph{noise}'', whereas local features gain more attention during evaluations of ``\emph{diagnosis}'' or ``\emph{lesions}''. 

Overall, the above findings indicate that our \modelname is capable of successfully performing the CT subjective quality assessment task. 
It can not only predict texts aligned with the ground truth but also translate these predictions into scores and reports through ChatGPT in a clinically meaningful way.

\section{Discussion}
Our study highlights the efficacy of integrating large models for image quality assessment, with a specific focus on low-dose CT denoising.
It suggests a significant potential to replace a traditional subjective image quality evaluation procedure conducted by radiologists with a large hybrid deep model, which would be 
resource-efficient and time-saving. 
In other words, our developed \modelname has made the first attempt along this direction, and
\modelname not only eases the burden on radiologists by automating CT image quality assessment but also promises to aid radiologists in refining the diagnostic performance.

Our method was developed on a CT-IQA dataset of 1,000 image-text pairs annotated by a professional radiologist. For this purpose, we leveraged a prompt template to transform quality scores into text descriptions.
Having fine-tuned an image quality captioning model on CT-IQA dataset, \modelname can generate quality descriptions for different CT scans.
The utilization of ChatGPT as an interactive interface facilitates user engagement, allowing for versatile outputs including quality scores and comprehensive reports.

Our experimental results demonstrate the efficacy of \modelname, which can generate quality descriptions steadily and convert them to scores and reports successfully.
Our quantitative evaluation, using metrics for image captioning, classification, and regression tasks, underscores its superior  performance.
In addition, our ablation study shows the effectiveness of incorporating LLMs in subjective CT IQA tasks because it can integrate the expertise of radiologists with the advanced capabilities of LLMs. 
Furthermore, LLM provides an interpretation of generated results using the quality captioning model.
Note that while CLIP-IQA also employs LLMs, its limitation to training one metric at a time with simple text prompts restricts its applicability in complex medical IQA scenarios, especially when assessing fine structures and small lesions.

However, our work also has some limitations, which open avenues for  enhancements. Currently, our study is based on a relatively small dataset annotated by a single radiologist. Expanding this dataset will likely enhance the accuracy and reliability of our model, potentially making \modelname a new standard for medical image quality assessment.
Additionally, more interactive elements between radiologists and the model would be beneficial such as tracking radiologists' eyes. Future work might include a text-guided denoising model, allowing radiologists to refine poor-quality images and improve radiology reports.

\section{Conclusion}
In conclusion, our study presents a pioneering exploration into CT subjective quality assessment, utilizing an innovative amalgamation of vision-language models and ChatGPT. 
We collect CT-IQA, an image-text dataset comprising pairs of CT scans with quality scores annotated by an experienced radiologist. 
We develop \modelname, fine-tuned on a vision-language model using the CT-IQA dataset, which can integrate with ChatGPT to generate both quality scores and detailed reports.
The extensive experimental results not only demonstrate the feasibility of \modelname but also highlight the effectiveness of LLMs, marking a significant potential in the field of subjective image quality assessment integrated with LLMs.

\end{document}